\documentclass[10pt,twocolumn,letterpaper]{article}

\usepackage{wacv}
\usepackage{times}
\usepackage{epsfig}
\usepackage{graphicx}
\usepackage{amsmath}
\usepackage{amssymb}
\usepackage{color}
\usepackage{enumerate}
\usepackage{enumitem}
\usepackage{array}
\usepackage{caption}
\usepackage{cite}
\usepackage[dvipsnames]{xcolor}
\usepackage[export]{adjustbox}
\usepackage{breqn}
\usepackage{adjustbox}
\usepackage{subcaption}

\usepackage{dblfloatfix}    

\def\wacvPaperID{151} 

\wacvfinalcopy 

\ifwacvfinal
\fi

\ifwacvfinal
\usepackage[breaklinks=true,bookmarks=false]{hyperref}
\else
\usepackage[pagebackref=true,breaklinks=true,colorlinks,bookmarks=false]{hyperref}
\fi

\ifwacvfinal
\fi

\begin{document}

\title{DeepCFL: Deep Contextual Features Learning from a Single Image}


\author{Indra Deep Mastan and Shanmuganathan Raman\\
Indian Institute of Technology Gandhinagar\\
Gandhinagar, Gujarat, India\\
{\tt\small \{indra.mastan, shanmuga\}@iitgn.ac.in}}

\maketitle
\thispagestyle{empty}  

\begin{abstract}
Recently, there is a vast interest in developing image feature learning methods that are independent of the training data, such as deep image prior \cite{Ulyanov2018CVPR}, InGAN \cite{shocher2018internal, shocher2019inGan}, SinGAN \cite{shaham2019singan}, and DCIL  \cite{deep2019dcil}. These methods perform various tasks, such as image restoration, image editing, and image synthesis. In this work, we proposed a new training data-independent framework, called Deep Contextual Features Learning (DeepCFL), to perform image synthesis and image restoration based on the semantics of the input image. The contextual features are simply the high dimensional vectors representing the semantics of the given image. DeepCFL is a single image GAN framework that learns the distribution of the context vectors from the input image. We show the performance of contextual learning in various challenging scenarios: outpainting, inpainting, and restoration of randomly removed pixels. DeepCFL is applicable when the input source image and the generated target image are not aligned. We illustrate image synthesis using DeepCFL for the task of image resizing. 
\end{abstract}

\section{Introduction}\label{sec: introduction}
Recently, there has been a remarkable success for image restoration and image synthesis methods that do not use training data \cite{Ulyanov2018CVPR, shocher2018zero, mastan2019multi, shocher2018internal, shaham2019singan, sidorov2019deep, deep2019dcil} . One of the major challenges for the deep feature learning methods above is the limited contextual understanding in the absence of feature learning from training samples \cite{mastan2019multi}. Contextual learning is mostly studied for image inpainting \cite{pathak2016context} and image transformation tasks \cite{mechrez2018contextual}, where many pairs of source and target images are used to learn the image context.

\begin{figure} \captionsetup[sub]{font=scriptsize,labelfont={bf,sf}} \captionsetup{font=small,labelfont={bf,sf}} \centering 
\begin{subfigure}{0.23\linewidth}\includegraphics[width=\linewidth]{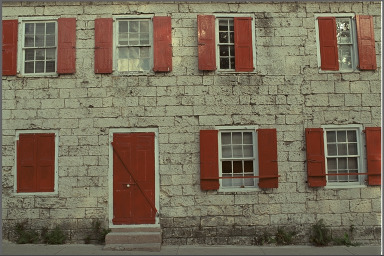}  \caption{Original} \end{subfigure}
\begin{subfigure}{0.23\linewidth}\includegraphics[width=\linewidth]{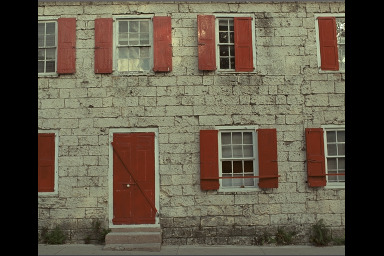}  \caption{Corrupted} \end{subfigure}
\begin{subfigure}{0.23\linewidth}\includegraphics[width=\linewidth]{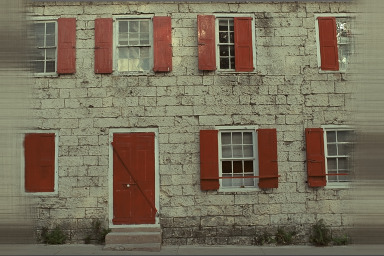} \caption{DIP \cite{Ulyanov2018CVPR}} \end{subfigure}
\begin{subfigure}{0.23\linewidth}\includegraphics[width=\linewidth]{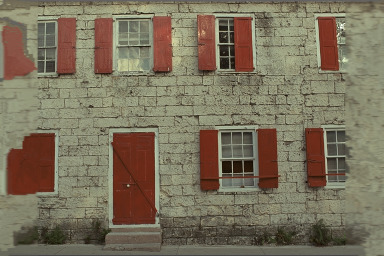} \caption{DeepCFL } \end{subfigure}   \\
\begin{subfigure}{0.9\linewidth}\includegraphics[width=\linewidth]{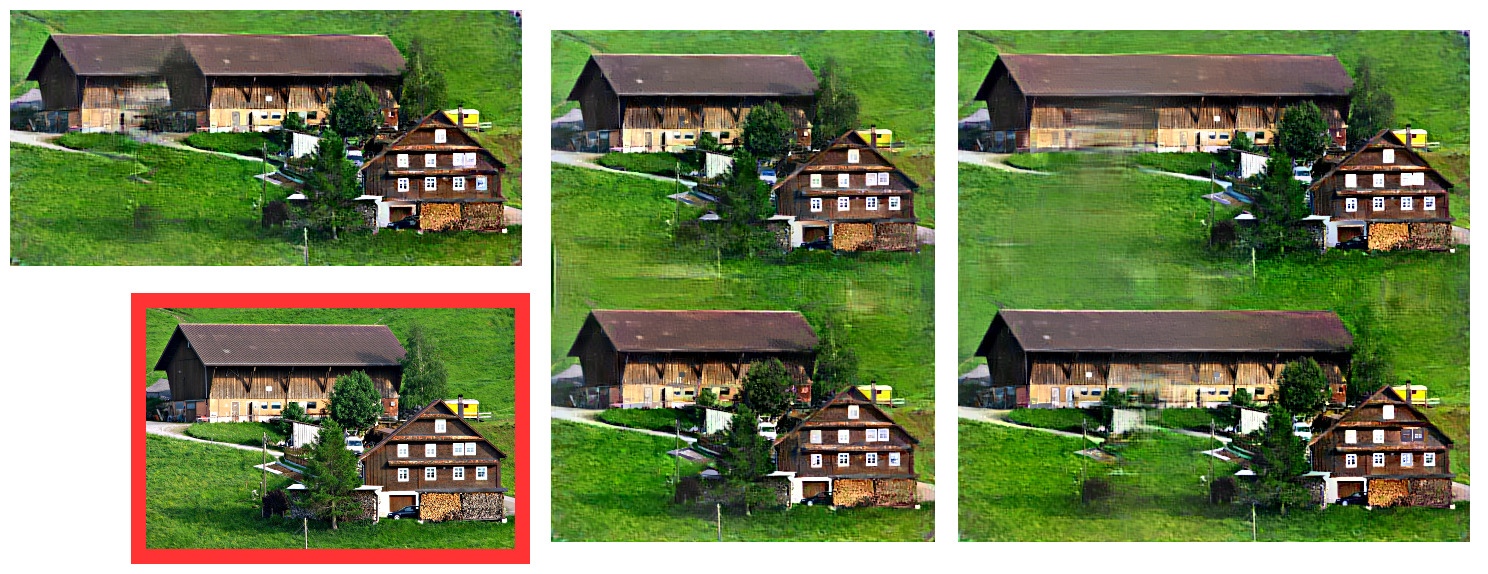}  \caption{Image synthesis using DeepCFL (input image in red color frame).} \end{subfigure}
\caption{The figure show image restoration (first row) and image synthesis (second row). Here, DIP \cite{Ulyanov2018CVPR} is a pixel-loss based setup. DeepCFL is a single image GAN framework for contextual learning. DeepCFL could fill the masked regions well for image restoration and also perform new object synthesis, which could not be performed using the pixel-based comparison of DIP \cite{Ulyanov2018CVPR}.}\label{fig: LOGO}
\end{figure}

\begin{figure*}\captionsetup{font=small,labelfont={bf,sf}} \centering
\begin{minipage}{0.6\linewidth} \includegraphics[width=\linewidth]{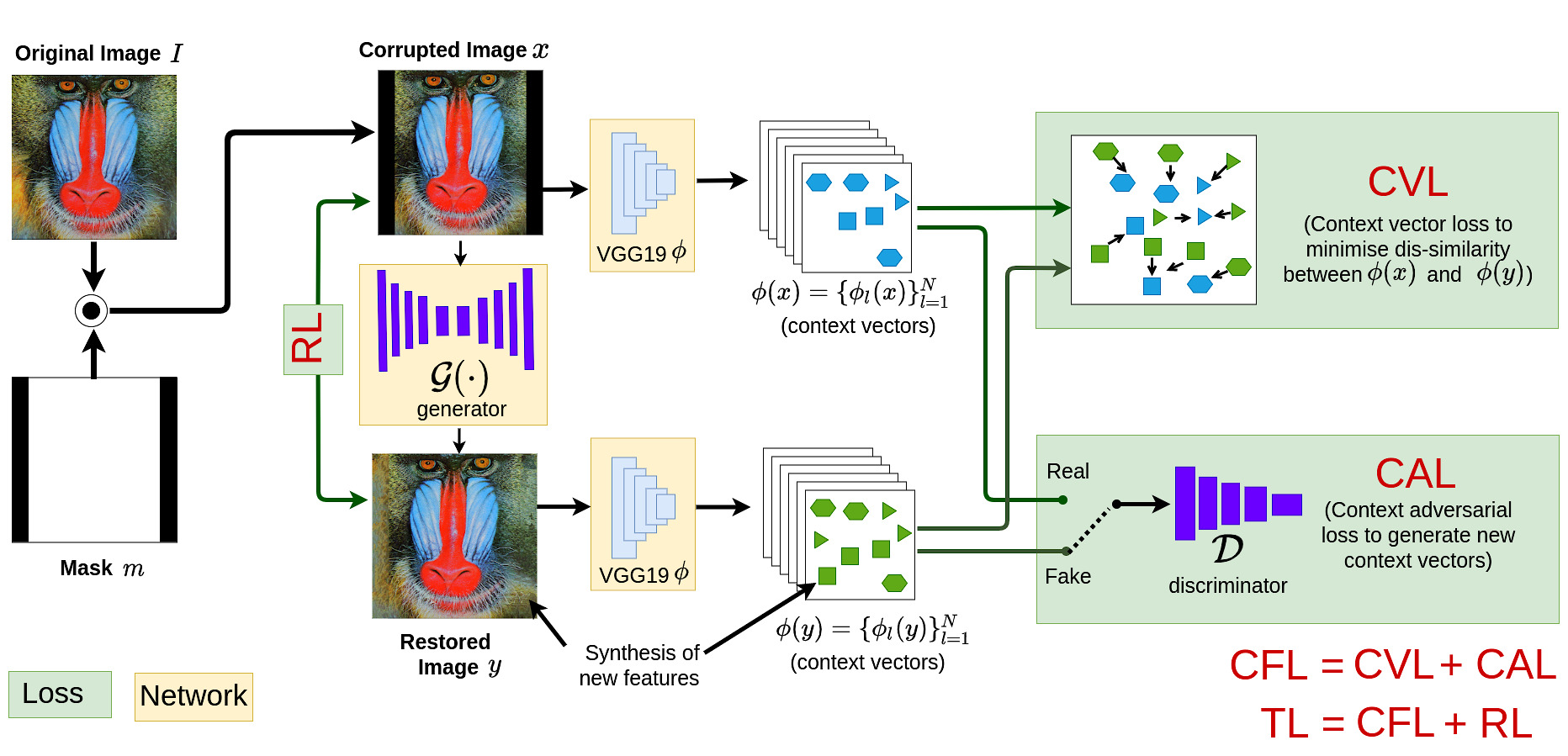}\caption{\textbf{Contextual Features Learning (DeepCFL).} The figure shows the framework for the outpainting. The corrupted image $x$ is fed into the generator $\mathcal{G}$. Here, $\mathcal{G}$ is an encoder-decoder network which outputs an image $\mathcal{G}(x)=y$. Next, VGG19 ($\phi$) computes the contextual features $\phi(x)$ of $x$ and $\phi(y)$ of $y$. Then we compute contextual features loss (CFL) and reconstruction loss (RL) and minimize total loss (TL). The main idea of DeepCFL is to synthesize new features by comparing image statistics at contextual features space. Note that CFL compares the context vectors $\phi(x)$ and $\phi(y)$ using CVL and CAL (see Fig.~\ref{fig: AblationContext} for above example). }\label{fig: mainModel} \end{minipage}\hspace*{0.5cm} %
\begin{minipage}{0.3\linewidth} \includegraphics[width=\linewidth]{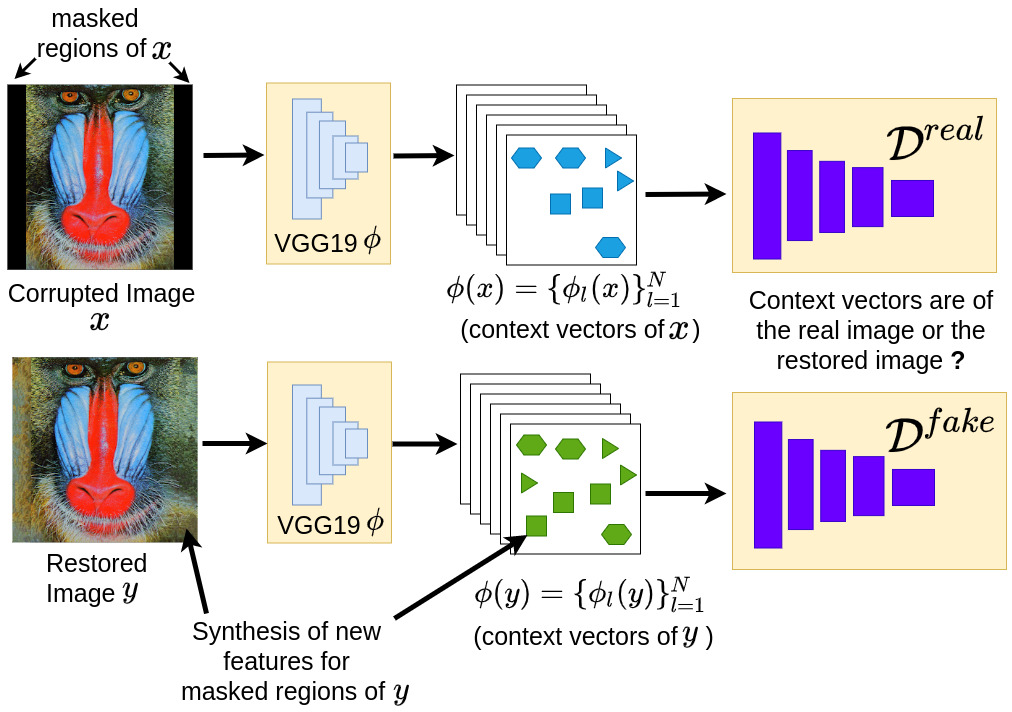}\caption{The figure shows the context vectors comparison in the adversarial framework. $\phi(x)$ and $\phi(y)$ are computed from the input image $x$ and the restored image $\mathcal{G}(x) = y$. The masked regions of $x$ are restored in $y$. For simplicity, we address restored regions in $y$ as masked regions of $y$. Here, $\mathcal{D}^{real}$ and $\mathcal{D}^{fake}$ are the two instances of  $\mathcal{D}$ which share the network parameters.}\label{fig: CALdiscriminator}\end{minipage}%
\end{figure*}

Restoration of missing pixels in an image is a classical inverse problem  \cite{zoran2011learning, levin2007blind, burger2012image, sun2008image, zhang2017image, buades2005non, dabov2007image, efros1999texture, simakov2008summarizing}.  It addresses various applications such as image editing, restoration of damaged paintings, image completion, and image outpainting. The image transformation model allows formulation for a variety of tasks such as style transfer, single image animation, and domain transfer \cite{mechrez2018contextual}. 

Traditionally, image restoration is formulated as optimization problems, where the objective function includes a loss term and an image prior term, \textit{e.g.}, sparse \cite{aharon2006k, dong2012nonlocally} and low-rank \cite{gu2017weighted} priors. The desired image is reconstructed by finding the solution for the optimization problem.  Deep learning models have shown an ability to capture image priors implicitly by minimizing the loss over the training samples \cite{ledig2016photo,  pathak2016context,  liu2018image, zhang2017learning, bigdeli2017image, yang2017high, wang2017high, wang2019wide, yu2018generative}. However, training data-based methods have their limitations, such as generalizability to new images \cite{Ulyanov2018CVPR,  mastan2019multi}.

Recently, there is a growing interest in developing methods that are independent of training data to perform image restoration and image synthesis tasks \cite{Ulyanov2018CVPR, shocher2018zero, mastan2019multi, shocher2018internal, shaham2019singan, sidorov2019deep, deep2019dcil}. Ulyanov \textit{et al.} proposed deep image prior (DIP) \cite{Ulyanov2018CVPR}, which shows that the handcrafted structure of the convolution neural network (CNN) provides an implicit image prior \cite{Ulyanov2018CVPR}. However, image prior learning using pixel-to-pixel loss in \cite{Ulyanov2018CVPR} is limited to the tasks which have a spatial correspondence between the pixels of the source image and the target image \cite{mechrez2018contextual}. One approach would be to learn the internal patch distribution from the input image when the source and the target images are not aligned.



The single image GAN frameworks show applications where the spatial mapping between the source and the target images is not well-defined \cite{shaham2019singan, shocher2018internal, shocher2019inGan, shocher2018zero}. Shocher \textit{et al.} proposed an internal learning (IL) framework to synthesize realistic image patches using image-conditional GAN, called InGAN \cite{shocher2018internal, shocher2019inGan}. Shaham \textit{et al.} showed an unconditional generative model for image synthesis, named SinGAN \cite{shaham2019singan}. Mastan \textit{et al.} have shown the single image GAN framework for denoising-super resolution and image resizing \cite{deep2019dcil}. 

The pixel-to-pixel loss framework in \cite{Ulyanov2018CVPR} and the internal patch distribution learning frameworks in \cite{shaham2019singan, shocher2018internal, shocher2019inGan} do not perform image reconstruction by considering the context of the objects. An image could be considered as a collection of high dimensional context vectors \cite{mechrez2018contextual}. These high dimensional vectors are the image statistics captured at the intermediate layers of the features extractor such as VGG19 network \cite{mechrez2018contextual, mechrez2018learning}. An interesting question would be that, given an incomplete image summary, can we synthesize new context vectors and use them to reconstruct the image. The context of an image is critical to perform image restoration and image synthesis tasks (Fig.~\ref{fig: LOGO} and Fig.~\ref{fig: retargetingGAN2}) \cite{Ulyanov2018CVPR, mastan2019multi, shocher2018zero, shocher2018internal}. We present a single image GAN framework (DeepCFL) which studies the contextual features in the image. The problem is \emph{novel} as it aims to learn the distribution of the contextual features (contextual learning) in the image instead of internal patch distribution, as in the case of InGAN \cite{shocher2018internal, shocher2019inGan} and SinGAN \cite{shaham2019singan}.

%
%

We have shown a pictorial representation of DeepCFL in Fig.~\ref{fig: mainModel}. The aim is to utilize the image features of the original image $I$, which are present in the corrupted image $x$. We generate a restored image $y$ which utilizes image features from $x$. We use an encoder-decoder network $\mathcal{G}$ to generate $y$. Then, we iteratively minimize the total loss (TL) between the corrupted image and the restored image. TL is a combination of contextual features loss (CFL) and reconstruction loss (RL). Fig.~\ref{fig: mainModel} shows that CFL allows feature learning using two different tools: contextual adversarial loss (CAL) and context vectors loss (CVL). The detailed description of each component of the framework and the formal definitions of the loss functions are described Sec.~\ref{sec: methodology}. 

CAL performs distribution matching in the adversarial framework to synthesize new context vectors for the corrupted image $x$. CVL computes the direct difference between the context vectors extracted from the corrupted image $x$ and the restored image $y$. Therefore, in CFL, CAL generates new context vectors and CVL improvises them. RL is a pixel-to-pixel loss (\textit{i.e.}, mean squared error), which ensures the preservation of image features in the restored images. Intuitively, the main idea is to generate new context vectors using CFL and map them to the image features implicitly through pixel-based comparison using RL.


We have studied the performance of DeepCFL for the following tasks: image outpainting, inpainting of arbitrary holes, and restoration of $r\%$ pixels missing in the corrupted image. We also show the applications in the presence of non-aligned image data using image resizing. The key contributions of this work are summarized below.

\begin{itemize}[leftmargin=*, nolistsep]
\item We propose a single image GAN framework for contextual features learning (DeepCFL). The framework performs well on image outpainting tasks (Fig.~\ref{fig: Outpainting} and Table~\ref{table: out}). We also illustrate that DeepCFL synthesizes new objects when resizing the image (Fig.~\ref{fig: retargetingGAN1}).  
\item DeepCFL investigates image reconstruction considering the contextual features. The contextual features learning is useful for the applications that use only a single image as input. We show the generalizability of DeepCFL by performing multiple applications (Sec.~\ref{sec: application}). 
\item We provide a detailed analysis of contextual features learning by illustrating reconstruction in various challenging setups such as arbitrary hole inpainting, restoration of a high degree of corruption, restoration of images with a word cloud, ablation studies, and limitations (Sec.~\ref{sec: application} and  Sec.~\ref{sec: ablationLimit}). 
\end{itemize}

%

\section{Related work}\label{sec: related}
Deep feature learning captures good image features by using the strong internal data repetitions (self-similarity prior) \cite{elad2006image, irani2009super, zontak2011internal, shocher2018zero, zhang2019internal}, hand-crafted structure \cite{Ulyanov2018CVPR, mastan2019multi}, and explicit regularizer \cite{mataev2019deepred}. DeepCFL is a single image GAN setup, which is different from features learning frameworks proposed earlier \cite{wang2019wide, teterwak2019boundless, yang2019very,  ren2019structureflow, liu2018image, yu2019free, nazeri2019edgeconnect, mechrez2018contextual}. Single image GAN frameworks performs variety of tasks such as image editing \cite{shaham2019singan}, retargeting \cite{shocher2019inGan}, denoising super-resolution \cite{deep2019dcil}, and video inpainting  \cite{zhang2019internal, kim2019deep}. Our contextual learning framework is somewhat related to \cite{Ulyanov2018CVPR, mastan2019multi, shocher2018internal, deep2019dcil}. InGAN \cite{shocher2019inGan, shocher2018internal} and SinGAN \cite{shaham2019singan} are single image GAN frameworks for learning the internal patch distribution. DCIL leverage internal learning with the contextual loss \cite{deep2019dcil}. DeepCFL is related to \cite{Ulyanov2018CVPR, mastan2019multi, shocher2018internal, deep2019dcil} and does not employ a masked patch discriminator for CAL \cite{wang2019wide}. It does not use a features expansion network and relies on the features reconstruction capabilities of the encoder-decoder network \cite{wang2019wide}.



\section{Our Framework} \label{sec: methodology}
DeepCFL is a single image GAN framework to synthesize new context vectors that are consistent with the semantics of the input source image. The task is to extract features from the source image and synthesize a new target image. The source image could be a clean or a corrupted image. The target image could be of the same size as the source image or a different size. For example, in the case of image restoration, we use a corrupted source image with missing pixel values. The contextual features are used to fill the missing regions of the corrupted image. For image synthesis, a clean image is used to synthesize new images of different sizes. Below, we discuss image restoration and context vectors before we describe the DeepCFL framework.

Let  $\mathcal{I}$ denote the set of original images,  $X$ denote the set of corrupted images, and $Y$ denote the set of restored images. Let $x$ denotes a corrupted image, \textit{i.e.}, $x\in X$. $x$ is computed by removing pixels from an original image $I$ using a binary mask $m$ as follows: $x = I \odot m$, where $\odot$ is the Hadamard product and $I\in \mathcal{I}$. The mask $m$ defines the underlying image restoration application. For example, in image outpainting of 20\% pixels, the mask removes the 10\% pixels each along the right side and the left side of the image. For the restoration of $r\%$ pixels, the mask contains $r\%$ zeros at random locations. For image inpainting, the mask contains arbitrary shapes. The objective is to restore the image details in $x$, which were removed by $m$.

\noindent \textbf{Image restoration procedure.} The task is to generate a new image $\mathcal{G}(x)=y$, which contains the restored pixels. Here, $\mathcal{G}$ is the generator network which maps the corrupted image to a restored image $y$, \textit{i.e.}, $y \in Y$. The corrupted image $x$ could be considered as a source image as it contains the features from the original image $I$.  The main intuition is to estimate the context for masked regions of $y$ based on the image features present at the unmasked regions of $x$ (Fig.~\ref{fig: CALdiscriminator}). The image restoration process iteratively minimizes the loss computed between $x$ and $y$. 

\noindent \textbf{What are context vectors?} The context vectors of an image $I$ are the image statistics present at intermediate layers of a feature extractor $\phi(I)$. VGG19 has been widely used to extract image statistics. Formally, given an image $I$, let $\phi(I) = \{ \phi_l(I) \}_{l=1}^{N}$ denote the set of context vectors extracted from $I$.  Here, $\phi : \mathcal{I} \rightarrow CV $ is  the pre-trained VGG19 network \cite{gatys2016image} which maps image $I \in \mathcal{I}$ to its context vectors $\phi(I) \in CV$. $\phi_l(\cdot)$ denotes the feature extracted from the layer $l$ of $\phi(\cdot)$ and $N$ is the number of layers in $\phi$. 

\noindent \textbf{Why context vectors are important?} 
Fig.~\ref{fig: LOGO} and Fig.~\ref{fig: retargetingGAN2} show that the contextual learning framework would allow image restoration and image synthesis based on the semantics of the input (refer Fig.~\ref{fig: Outpainting} and Fig.~\ref{fig: retargetingGAN1} for more examples).  For example, in the case of restoration of missing pixels, the key observation is to improve the masked regions in the restore image $y$ using the unmasked regions in the corrupted image $x$. It is done by matching the distribution of the contextual features of the corrupted image $\phi(x)$ and the contextual features of the restored image $\phi(y)$ (Sec.~\ref{ssec: lossFn}).

\noindent \textbf{DeepCFL.} We now discuss the DeepCFL framework shown in Fig.~\ref{fig: mainModel}.  It consists of a generator $\mathcal{G}$, a discriminator $\mathcal{D}$, and a features extractor $\phi$. The corrupted image $x$ is fed into $\mathcal{G}$. The generator outputs an image $y=\mathcal{G}(x)$. Next, we feed $x$ and $y$ into $\phi(\cdot)$ to compute $\phi(x)$ and $\phi(y)$. Then we minimize the total loss (TL) computed between $x$ and $y$ (Eq.~\ref{eq: inpaintingL}). The two primary components of TL are the contextual features loss (CFL) and the reconstruction loss (RL). CFL synthesizes new context vectors for the masked regions in $\phi(y)$, where the features learning procedure is assisted by contextual features in $\phi(x)$. $\mathcal{D}$ is used for computing CFL. RL is computed between the unmasked regions of $x$ and $y$ to provide image feature consistency in $y$.

\subsection{Network Design} \label{ssec: networkDesign}

\noindent \textbf{Generator.}  The generator $\mathcal{G}: X \rightarrow Y$ maps the source image $x \in X$ to the target image $y \in Y$. $\mathcal{G}$ is a depth-5 encoder-decoder network without skip connections (ED). The ED architecture works as the implicit regularizer to stabilize the image feature learning \cite{mastan2019multi, Ulyanov2018CVPR}. It exploits the inherent self-similarity present in the source image. We use context normalization \cite{wang2019wide} to maximize features learning. Intuitively, DeepCFL is unsupervised in the sense that no training data are used to train the generator network for any of the tasks. It is a single image GAN framework which uses pre-trained VGG19 as the features extractor. VGG19 is widely used in style transfer works for defining loss at VGG features space. The feature extractor distills strong prior in the framework \cite{deep2019dcil}. 

\noindent \textbf{Discriminator.} The discriminator $\mathcal{D}: CV \rightarrow \mathcal{M}$ maps the context vectors to a discriminator map $\mu \in \mathcal{M}$, where each entry in $\mu$ denotes the probability of the context vector coming from the distribution of the contextual feature of the original image.  Fig.~\ref{fig: CALdiscriminator} illustrates the discriminator task to distinguish context vectors $\phi(x)$ and $\phi(y)$. The generator $\mathcal{G}$ learns the context vectors through its interaction with $\mathcal{D}$. We use a multi-scale discriminator (MSD), where each output is a weighted average of the output from several discriminators (we have illustrated $\mathcal{D}$ using a single CNN for simplicity in Fig.~\ref{fig: mainModel} and Fig.~\ref{fig: CALdiscriminator}). Note that the discriminators in MSD would resize the context vectors.



\subsection{Loss Function}  \label{ssec: lossFn} 
\begin{figure*}[!ht]
\captionsetup[sub]{font=small,labelfont={bf,sf}}\captionsetup{font=small,labelfont={bf,sf}} \centering 
\begin{subfigure}[b]{0.140\textwidth} \includegraphics[width=\linewidth]{images/outpainting/new/kodim01.jpg} \end{subfigure}
\begin{subfigure}[b]{0.140\textwidth} \includegraphics[width=\linewidth]{images/outpainting/new/kodim01_mask.jpg} \end{subfigure}
\begin{subfigure}[b]{0.140\textwidth} \includegraphics[width=\linewidth]{images/outpainting/new/DIP_out__kodim01_.jpg} \end{subfigure}
\begin{subfigure}[b]{0.140\textwidth} \includegraphics[width=\linewidth]{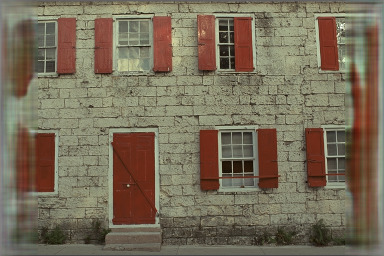} \end{subfigure}
\begin{subfigure}[b]{0.140\textwidth} \includegraphics[width=\linewidth]{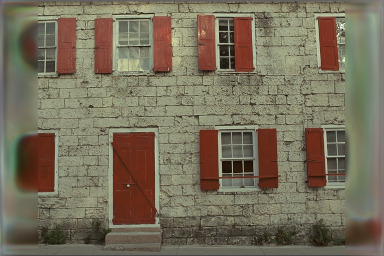} \end{subfigure}
\begin{subfigure}[b]{0.140\textwidth} \includegraphics[width=\linewidth]{images/outpainting/new/kodim01_DCIL_CInGAN_20_P.jpg} \end{subfigure}
\begin{subfigure}[b]{0.140\textwidth} \includegraphics[width=\linewidth]{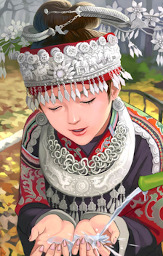} \end{subfigure}
\begin{subfigure}[b]{0.140\textwidth} \includegraphics[width=\linewidth]{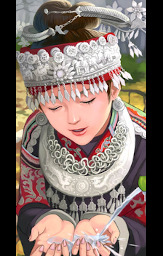} \end{subfigure}
\begin{subfigure}[b]{0.140\textwidth} \includegraphics[width=\linewidth]{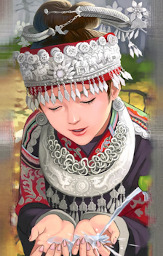} \end{subfigure}
\begin{subfigure}[b]{0.140\textwidth} \includegraphics[width=\linewidth]{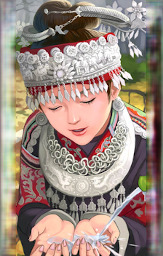} \end{subfigure}
\begin{subfigure}[b]{0.140\textwidth} \includegraphics[width=\linewidth]{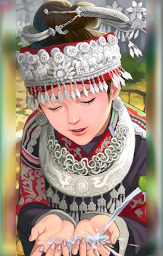} \end{subfigure}
\begin{subfigure}[b]{0.140\textwidth} \includegraphics[width=\linewidth]{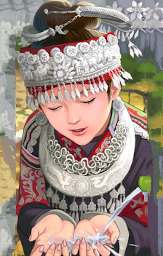} \end{subfigure}
\begin{subfigure}[b]{0.140\textwidth} \includegraphics[width=\linewidth]{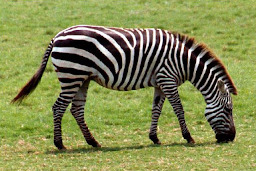} \end{subfigure}
\begin{subfigure}[b]{0.140\textwidth} \includegraphics[width=\linewidth]{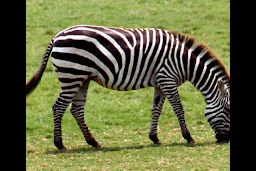} \end{subfigure}
\begin{subfigure}[b]{0.140\textwidth} \includegraphics[width=\linewidth]{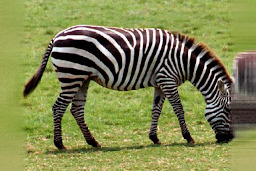} \end{subfigure}
\begin{subfigure}[b]{0.140\textwidth} \includegraphics[width=\linewidth]{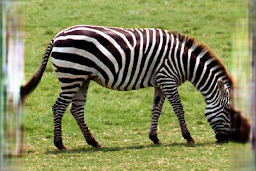} \end{subfigure}
\begin{subfigure}[b]{0.140\textwidth} \includegraphics[width=\linewidth]{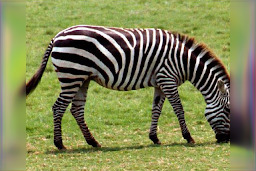} \end{subfigure}
\begin{subfigure}[b]{0.140\textwidth} \includegraphics[width=\linewidth]{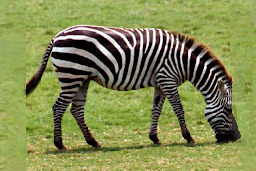} \end{subfigure}
\begin{subfigure}[b]{0.140\textwidth} \includegraphics[width=\linewidth]{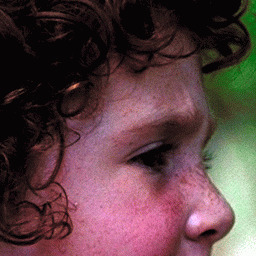} \caption{Original}  \end{subfigure}
\begin{subfigure}[b]{0.140\textwidth} \includegraphics[width=\linewidth]{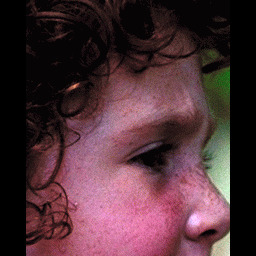}  \caption{Corrupted}  \end{subfigure}
\begin{subfigure}[b]{0.140\textwidth} \includegraphics[width=\linewidth]{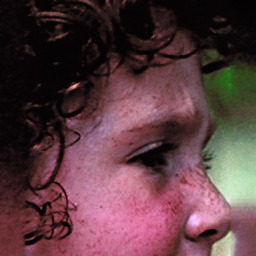}  \caption{DIP \cite{Ulyanov2018CVPR} } \end{subfigure}
\begin{subfigure}[b]{0.140\textwidth} \includegraphics[width=\linewidth]{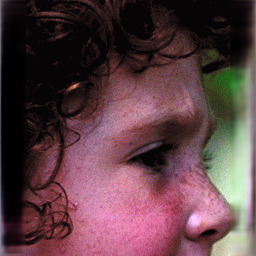} \caption{MEDS \cite{mastan2019multi} } \end{subfigure}
\begin{subfigure}[b]{0.140\textwidth} \includegraphics[width=\linewidth]{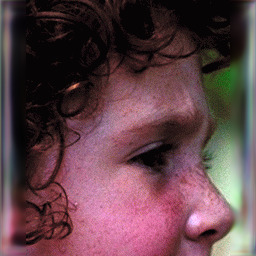} \caption{InGAN \cite{shocher2018internal} } \end{subfigure}
\begin{subfigure}[b]{0.140\textwidth} \includegraphics[width=\linewidth]{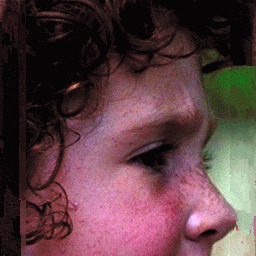} \caption{DeepCFL} \end{subfigure}
\caption{\textbf{Image outpainting.} The figure shows the restoration of 20\% pixels in the image. DIP \cite{Ulyanov2018CVPR} and MEDS \cite{mastan2019multi} fill the missing regions but do not preserve the structure of the objects. Internal learning of InGAN \cite{shocher2018internal} performed better, but the generated new image features are not very clear. DeepCFL incorporates the contextual understanding and is observed to perform better (Table~\ref{table: out}).  }
\label{fig: Outpainting}
\end{figure*} 
The goal of the loss function is to maximize the feature learning from source $x$ by comparing it with generated image $\mathcal{G}(x)=y$. The total loss (TL) is defined in Eq.~\ref{eq: inpaintingL}. 
\begin{equation}\label{eq: inpaintingL}
\begin{split}
\mathcal{L}_{tl}(x, y, \mathcal{G}, \mathcal{D}, \phi) = \lambda_{\mathcal{G}} \; \mathcal{L}_{cfl}(x, y, \mathcal{G}, \mathcal{D}, \phi) \\ + \lambda_{\mathcal{R}} \; \mathcal{L}_{rl}(x, y, \mathcal{G})
\end{split}
\end{equation}
Here, $\mathcal{L}_{cfl}$ denotes CFL and $\mathcal{L}_{rl}$ denotes RL. The terms $\lambda_{\mathcal{G}}$ and $\lambda_{\mathcal{R}}$ are the coefficients of CFL and RL. We have pictorially shown CFL and RL in Fig.~\ref{fig: mainModel}.  
The total loss described in Eq.~\ref{eq: inpaintingL} compares the image features in two ways: CFL and RL. CFL provides new image features to $y$, which are consistent with the object context of $x$. RL maximizes the likelihood of randomly initialized network weights. 

\subsubsection{Contextual Features Loss (CFL)} The purpose of CFL is to learn the distribution of context vectors to synthesize image features in $y$ based on the semantics of the input $x$. We extract context vectors $\phi(x)$ and $\phi(y)$ and then minimize the loss described in Eq.~\ref{eq: cfl}. 
\begin{equation}\label{eq: cfl}
\begin{split}
\mathcal{L}_{cfl}(x, y, \mathcal{G}, \mathcal{D}, \phi) = \lambda_{cal} & \mathcal{L}_{cal}(\mathcal{G}, \mathcal{D};\phi) \\ & + \lambda_{cvl} \mathcal{L}_{cvl}(\phi(x), \phi(y))
\end{split}
\end{equation}
Here, $\mathcal{L}_{cfl}$ denotes CFL, $\mathcal{L}_{cal}$ denotes CAL, and $\mathcal{L}_{cvl}$ denotes CVL. $\lambda_{cal}$ and $\lambda_{cvl}$ are the coefficients of CAL and CVL. 
Eq.~\ref{eq: cfl} shows that CFL compares the context vectors in two ways. (1) Context vector comparison in the adversarial framework using CAL. (2) Contextual features comparison by computing cosine distance in CVL.  CAL is an adversarial loss computed using the generator $\mathcal{G}$ and the discriminator $\mathcal{D}$. It is aimed to synthesize new contextual features that are indistinguishable from the features of the source image. The CVL computes the difference between contextually similar vectors to make the synthesized features of $y$ similar to the features of $x$.

\noindent \textbf{Context Adversarial Loss (CAL).} We have used the LSGAN \cite{mao2017least} variant of the adversarial learning framework. 
\begin{equation}
\mathcal{G}^* = \min_{\mathcal{G}} \max_{\mathcal{D}} \mathcal{L}_{cal}(\mathcal{G}, \mathcal{D}; \phi)
\end{equation}
Here, $\mathcal{G}^{*}$ is the generator with optimal parameters. The loss $\mathcal{L}_{cal}$ is defined in Eq.~\ref{eq: cal}. 
\begin{equation}\label{eq: cal}
\begin{split}
\mathcal{L}_{cal}(\mathcal{G}, \mathcal{D}; \phi) = & \mathbb{E}_{x \sim pdata(x)} [ (\mathcal{D}(\phi(x))-1)^2] \\ &+ \mathbb{E}_{x \sim pdata(x)} [\mathcal{D}(\phi( \mathcal{G}(x)))^2 ] 
\end{split}
\end{equation}

Eq.~\ref{eq: cal} shows the distribution matching of context vectors of the restored image $\phi(y) = \phi (\mathcal{G}(x))$ and context vectors of the corrupted image $\phi (x)$. The discriminator $\mathcal{D}$ tries to determine whether the context vectors are from $x$ or $y$ (see Fig.~\ref{fig: mainModel} and Fig.~\ref{fig: CALdiscriminator}).  Intuitively, this would help us to fill the context of the masked regions of $y=\mathcal{G}(x)$ by learning the context of the objects in unmasked areas in $x$. We have described $\mathcal{G}$, $\mathcal{D}$, and $\phi$ in Sec.~\ref{ssec: networkDesign}. %

\noindent \textbf{Context Vector Loss (CVL).} The main purpose of CVL is to improve the quality of contextual features in  $\phi(y)$ learned by CAL. $\mathcal{L}_{cvl}(\phi_l(x), \phi_l(y))$ is the sum of the contextual loss \cite{mechrez2018contextual} computed at each layer $l$ in $\phi$.  We have defined CVL for layer $l$ in Eq.~\ref{eq: cl}.
\begin{equation}\label{eq: cl} 
\mathcal{L}_{cvl}(\phi_l(x), \phi_l(y), l) = - \log (CX(\phi_l(x), \phi_l(y)))
\end{equation}
Here, $CX$ is the contextual similarity defined using the cosine distance between the features contained in $\phi_l(x)$ and  $\phi_l(y)$.  Note that $CX$ is computed by finding for each feature $\phi_l(y)_j$, a feature $\phi_l(x)_i$  that is most similar to it and then summed for all $\phi_l(y)_j$. Fig.~\ref{fig: mainModel} illustrate the matched context vectors of $\phi_l(x)_i$ and $\phi_l(y)_j$ by an arrow. Intuitively, the feature matching performed between the context vectors of masked regions of $y$ and the context vectors of unmasked regions of $x$ enables feature refinements for the new context vectors created by CAL. We  used $conv4\_2$ layer of $\phi$ to compute context vectors as the higher layers capture the high-level content in terms of objects structure \cite{gatys2016image}. It is interesting to note that CVL is different from perceptual loss $\|\phi_l(x) - \phi_l(y)\|$, which computes features difference without using contextual similarity criterion.

\subsubsection{Reconstruction Loss (RL).} 
RL is aimed to preserve image features and it is computed between corrupted  image $x$ and restored image $\mathcal{G}(x)=y$ (Fig.~\ref{fig: mainModel}). Let $\mathcal{L}_{rl}$ denotes RL. We define $\mathcal{L}_{rl}$ in Eq.~\ref{eq: reconstructionLoss}.
\begin{equation}\label{eq: reconstructionLoss}
\mathcal{L}_{rl}(\mathcal{G}, x, y) = \|\mathcal{G}(x) \odot m - x  \| 
\end{equation}
Eq.~\ref{eq: reconstructionLoss} shows the comparison between unmasked regions of $x$ with the unmasked regions of $y$. The unmasked regions in $x$ contains image features from $I$ and masked regions in $x$ are corrupted due to mask, \textit{i.e.}, $x=I \odot m $.  RL is a pixel-wise loss and it imposes a strong self-similarity prior \cite{Ulyanov2018CVPR}.

\begin{figure*}[!ht] \captionsetup[sub]{font=small,labelfont={bf,sf}}  \captionsetup{font=small,labelfont={bf,sf}} \centering  \resizebox{0.88\linewidth}{!}{ 
\begin{subfigure}{0.14\linewidth} \captionsetup{justification=centering} \includegraphics[width=\linewidth]{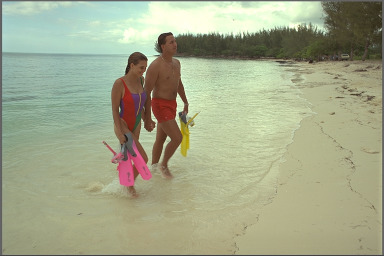} \caption{ Original image} \end{subfigure}
\begin{subfigure}{0.14\linewidth} \captionsetup{justification=centering} \includegraphics[width=\linewidth]{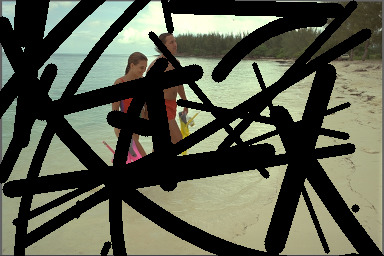}  \caption{ Corrupted} \end{subfigure}
\begin{subfigure}{0.14\linewidth} \captionsetup{justification=centering} \includegraphics[width=\linewidth]{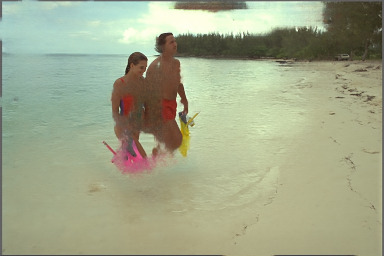}  \caption{ DIP, 0.90}  \end{subfigure}
\begin{subfigure}{0.14\linewidth} \captionsetup{justification=centering} \includegraphics[width=\linewidth]{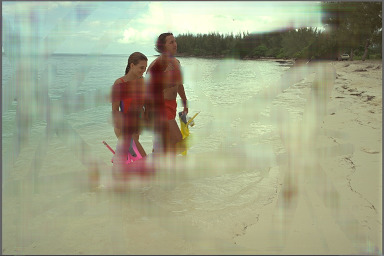}  \caption{MEDS, 0.89}  \end{subfigure}
\begin{subfigure}{0.14\linewidth}  \captionsetup{justification=centering} \includegraphics[width=\linewidth]{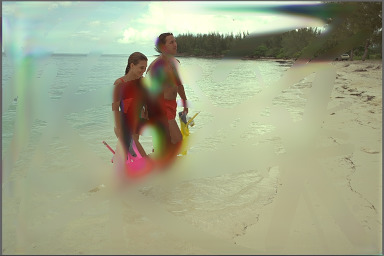}  \caption{InGAN, 0.90}  \end{subfigure}
\begin{subfigure}{0.14\linewidth} \captionsetup{justification=centering} \includegraphics[width=\linewidth]{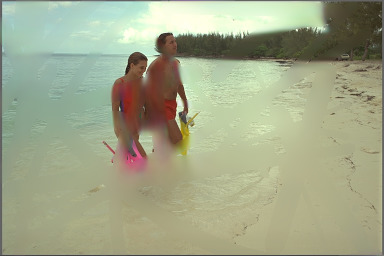}  \caption{DeepCFL, 0.91}  \end{subfigure}} 
\caption{ \textbf{Inpainting.} The figure shows the inpainting of arbitrary holes for DIP \cite{Ulyanov2018CVPR}, MEDS \cite{mastan2019multi}, InGAN \cite{shocher2018internal}, and DeepCFL (ours). DeepCFL minimize the features spillover of trees and attains better perceptual quality. }
\label{fig: inpainting}
\end{figure*}%

\section{Applications}\label{sec: application}
Here, we discuss the following applications of DeepCFL. (1) Image outpainting: extension of an image along the sides. (2) Image inpainting of irregular holes in the image. (3) Content-aware image resizing: synthesis of new objects when we resize an image. (4) Restoration in the presence of high degree of corruption: $50\%$ pixels\footnote{We have used original implementations of DIP \cite{Ulyanov2018CVPR}, MEDS \cite{mastan2019multi}, and DCIL \cite{deep2019dcil}. We implemented image restoration using the internal learning of InGAN \cite{shocher2018internal}. We have provided the implementation details in the supplementary material.}.

\subsection{Image Outpainting. } \label{ssec: outpainting}
Image outpainting relates to image extension, which creates new features while maintaining the semantics of the scene. Image extension uses training data to learn image context and then generates the complete scene given partial information \cite{wang2019wide, teterwak2019boundless, yang2019very, pathak2016context}. Our outpainting task does not use any training samples and synthesize features using only the corrupted image. We address outpainting as an image extension for convenience. 

A good image outpainting approach would fill the image features based on the semantics of the object present at the boundaries.  The ability of the generator to synthesize new contextual features over a large spatial extent along the sides depends upon the contextual learning. Unlike pixel-to-pixel loss, the context vectors based loss functions CFL (Eq.~\ref{eq: cfl}) aims to fill new features in the masked regions of the restored image, which are semantically similar to the unmasked regions of the corrupted image (refer Sec.~\ref{sec: methodology}).

In Fig.~\ref{fig: Outpainting}, we show outpainting of 20\% missing pixels, where the corrupted image is generated by removing 10\% pixels along the right side and the left side. DIP \cite{Ulyanov2018CVPR}, MEDS \cite{mastan2019multi}, and InGAN \cite{shocher2018internal} are contextual features learning independent methods. Image outpainting is better achieved using the semantics of the objects in the contextual learning-based DeepCFL framework.  
Table~\ref{table: out} shows the quantitative comparison on the standard datasets from \cite{heide2015fast}, Set5 and Set14 datasets \cite{mastan2019multi}. It could be observed that DeepCFL outperforms the other methods for outpainting. We have provided more details in the supplementary material.

\begin{table}[!ht]\setlength \extrarowheight{2pt} \captionsetup[sub]{font=small,labelfont={bf,sf}} \captionsetup{font=small,labelfont={bf,sf}}\centering
\resizebox{0.92\linewidth}{!}{%
\begin{tabular}{| p{20pt} | c | c | c | c |}\hline
	&	\textbf{DIP} \cite{Ulyanov2018CVPR}	&	\textbf{MEDS} \cite{mastan2019multi}	&	\textbf{InGAN} \cite{shocher2018internal}	&	\textbf{DeepCFL}\\\hline
SD    &     \begin{tabular}[c]{@{}l@{}}0.91\\ 23.73 \end{tabular}    &   \begin{tabular}[c]{@{}l@{}}0.91\\ 21.70\end{tabular}    &   \begin{tabular}[c]{@{}l@{}}\textbf{\textbf{0.92}}\\ 22.89\end{tabular}    &   \begin{tabular}[c]{@{}l@{}}\textbf{0.92}\\ \textbf{24.13}\end{tabular}\\\hline
Set14    &    \begin{tabular}[c]{@{}l@{}}0.89\\ 22.12\end{tabular}    &   \begin{tabular}[c]{@{}l@{}}0.89\\ 20.24\end{tabular}    &   \begin{tabular}[c]{@{}l@{}}\textbf{0.90}\\ 21.19\end{tabular}    &    \begin{tabular}[c]{@{}l@{}}\textbf{0.90}\\ \textbf{22.52}\end{tabular} \\\hline
Set5    &    \begin{tabular}[c]{@{}l@{}}0.88\\ 19.03 \end{tabular}    &   \begin{tabular}[c]{@{}l@{}}0.88\\ 19.35\end{tabular}    &   \begin{tabular}[c]{@{}l@{}}0.89\\ 19.29\end{tabular}    &    \begin{tabular}[c]{@{}l@{}}\textbf{0.90}\\ \textbf{21.50}\end{tabular}\\\hline
\end{tabular}} 
\caption{Quantitative comparision using SSIM values (top) and PSNR values (bottom) for image outpainting of 20\% pixels on standard dataset (SD), Set5 and Set14 datasets.  }\label{tab: outpainting} 
\label{table: out}%
\end{table}

\begin{figure*}\centering \captionsetup{font=small,labelfont={bf,sf}} \captionsetup[sub]{font=small,labelfont={bf,sf}}
\includegraphics[width=0.95\linewidth]{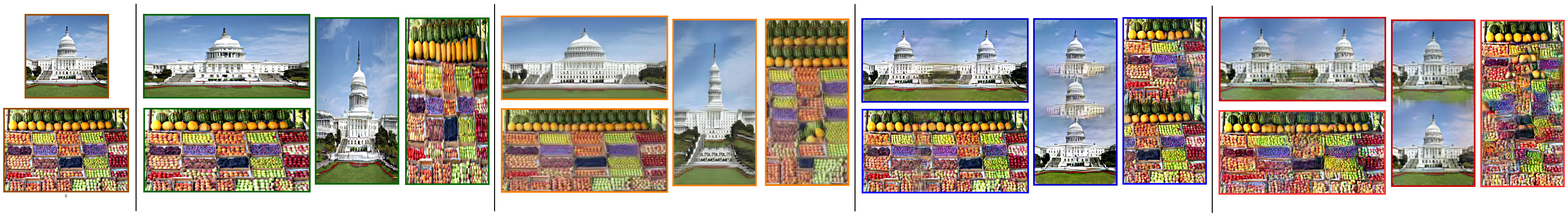}\vspace*{-0.2cm}
\hspace*{10pt}\begin{minipage}{0.07\linewidth}\centering (a) Input \end{minipage}%
\begin{minipage}{0.227\linewidth}\centering (b) Seam Carving \cite{avidan2007seam} \end{minipage}%
\begin{minipage}{0.225\linewidth}\centering (c) InGAN \cite{shocher2018internal} \end{minipage}%
\begin{minipage}{0.225\linewidth}\centering (d) DCIL \cite{deep2019dcil} \end{minipage}%
\begin{minipage}{0.225\linewidth}\centering (e) DeepCFL (ours)\end{minipage}%
\caption{\textbf{Image Resize.} The figure shows the synthesis of small objects (fruits) and large objects (building). Seam Carving (SC) \cite{avidan2007seam} does not preserve the structure well when resizing. For example, the shape of the fruits in small object synthesis is deformed in SC output. InGAN \cite{shocher2018internal} preserve the structure for small objects but does not preserve for large object (building). DCIL  \cite{deep2019dcil} synthesizes new objects when resizing. For example, object structure is preserved well when scaling $2\times$ along the width of the building. DeepCFL also preserves object structure when synthesizing new objects. It could be observed that DeepCFL does not duplicate the objects along the expended dimension. For example, DeepCFL synthesizes the fruits when resizing (the images are best viewed after zooming).} \label{fig: retargetingGAN1}
\end{figure*}
\begin{figure*}[!h] \captionsetup[sub]{font=small,labelfont={bf,sf}} \captionsetup{font=small,labelfont={bf,sf}} \centering \resizebox{0.94\linewidth}{!}{ 
\begin{subfigure}{0.145\linewidth} \captionsetup{justification=centering} \begin{center}
\includegraphics[width=\textwidth]{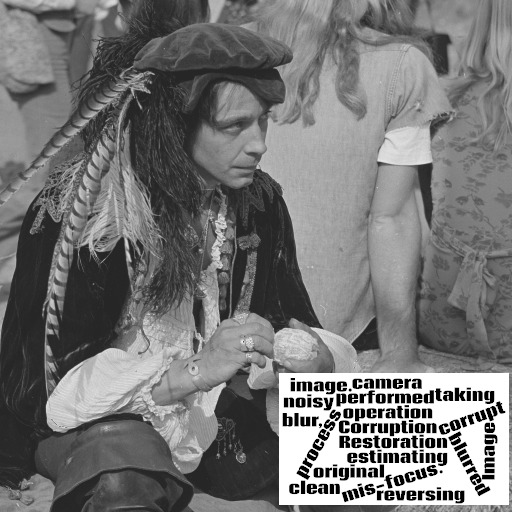} 
\end{center} \caption{Original} \end{subfigure}
\begin{subfigure}{0.145\linewidth} \captionsetup{justification=centering} \begin{center}
\includegraphics[width=\textwidth]{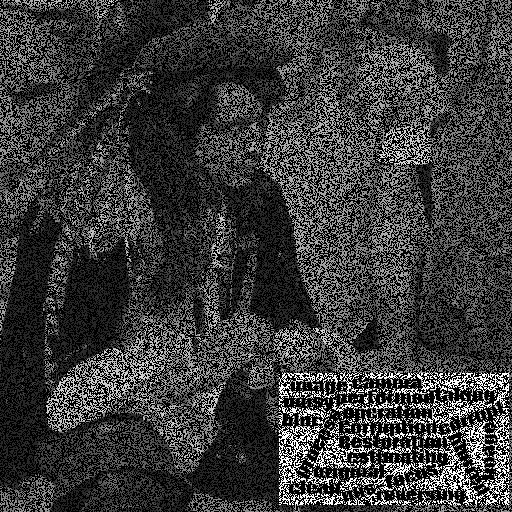} 
\end{center} \caption{Corrupted} \end{subfigure}
\begin{subfigure}{0.145\linewidth} \captionsetup{justification=centering} \begin{center}
\includegraphics[width=\textwidth]{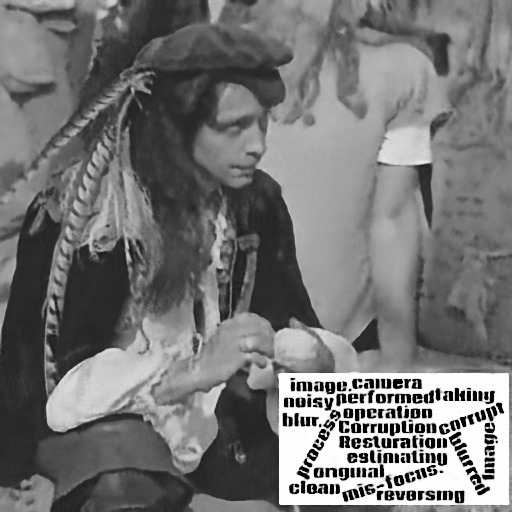}
\end{center} \caption{ DIP, 0.87 } \end{subfigure} 
\begin{subfigure}{0.145\linewidth} \captionsetup{justification=centering} \begin{center}
\includegraphics[width=\textwidth]{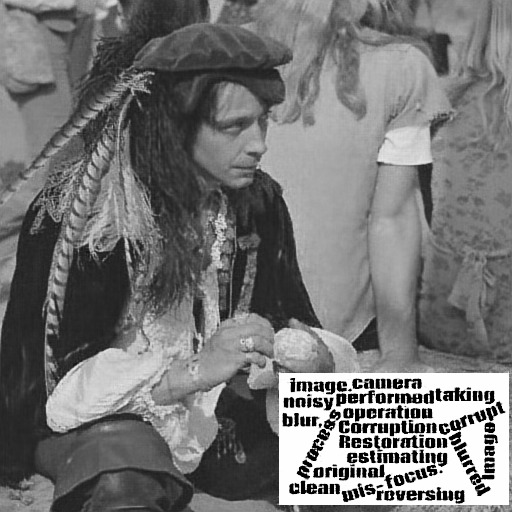}
\end{center} \caption{ MEDS, 0.92} \end{subfigure} 
\begin{subfigure}{0.145\linewidth}\captionsetup{justification=centering}\begin{center}
\includegraphics[width=\textwidth]{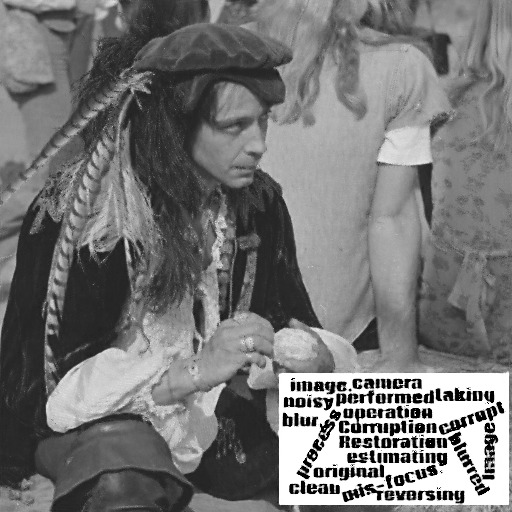}
\end{center} \caption{ InGAN, 0.93 } \end{subfigure} 
\begin{subfigure}{0.145\linewidth} \captionsetup{justification=centering} \begin{center}
\includegraphics[width=\textwidth]{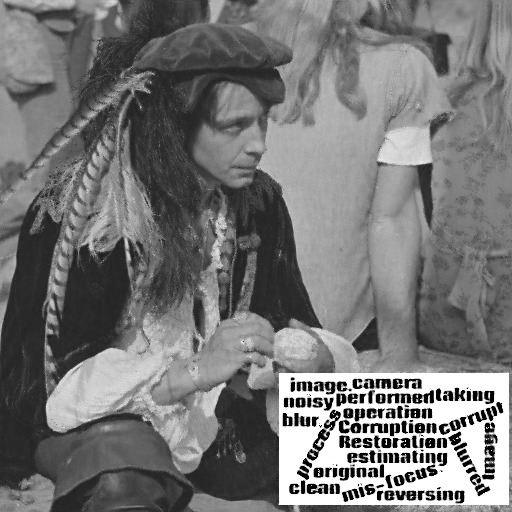}
\end{center} \caption{ DeepCFL, 0.93} \end{subfigure} } \vspace*{-0.2cm} 
\caption{ \textbf{RestoreWC 50\%.} The figure shows restoration in the presence of a word cloud. DeepCFL restore image features details comparable to DIP \cite{Ulyanov2018CVPR}, MEDS \cite{mastan2019multi}, and InGAN \cite{shocher2018internal}. 
}
\label{fig: inpaintingWd}
\end{figure*}

\subsection{Image Inpainting.}  \label{ssec: inpainting}
The input image has non-uniform corrupted regions spread across the entire image in the inpainting task.  It is a natural way by which an image could get corrupted \cite{liu2018image, ren2019structureflow}. The critical property to perform inpainting without using training data is to utilize the internal self-similarity property of the natural images \cite{Ulyanov2018CVPR, zhang2019internal}. The computation of the MSE between the generator output and the corrupted image tends to capture strong self-similarity prior \cite{Ulyanov2018CVPR}. DeepCFL leverages this learning by incorporating the context vectors comparison. The features learning procedure for inpainting is similar to outpainting described in Sec~\ref{ssec: outpainting}.

Fig.~\ref{fig: inpainting} shows the visual results for arbitrary hole inpainting. It could be observed that the contextual learning of DeepCFL minimizes the features spillover between different objects and fill the arbitrary holes considering the semantics of the image. The quantitative comparison (SSIM) for inpainting is as follows: DIP \cite{Ulyanov2018CVPR}: 0.90, MEDS \cite{mastan2019multi}: 0.88, InGAN \cite{shocher2019inGan} 0.90, and DeepCFL (ours): 0.91. We have provided more comparisons of generated images in the supplementary material. DeepCFL performs comparably to other frameworks. The estimation of the parameters from a single image is highly sensitive to the hyper-parameters (\textit{e.g.}, learning rate) \cite{Ulyanov2018CVPR, mastan2019multi}. We believe that the restoration quality of our method and other methods could be improved further using the hyper-parameter search.

\begin{figure}[!h]\captionsetup[sub]{font=scriptsize,labelfont={bf,sf}} \captionsetup{font=small,labelfont={bf,sf}}
\begin{center}
\resizebox{\linewidth}{!}{%
\begin{subfigure}[b]{0.15\linewidth}
\includegraphics[width=\linewidth, height=3cm,keepaspectratio]{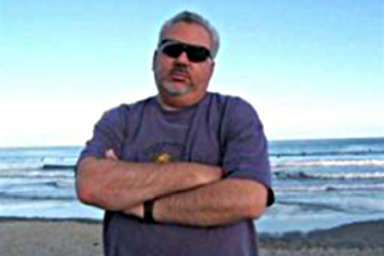}\caption{Input}
\end{subfigure}
\begin{subfigure}[b]{0.2\linewidth}
\includegraphics[width=\linewidth, height=3cm,keepaspectratio]{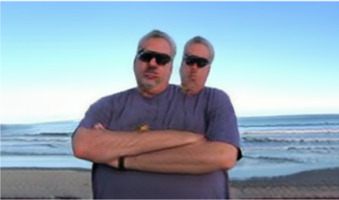}\caption{InGAN \cite{shocher2019inGan} }
\end{subfigure}
\begin{subfigure}[b]{0.26\linewidth}
\includegraphics[width=\linewidth, height=3cm,keepaspectratio]{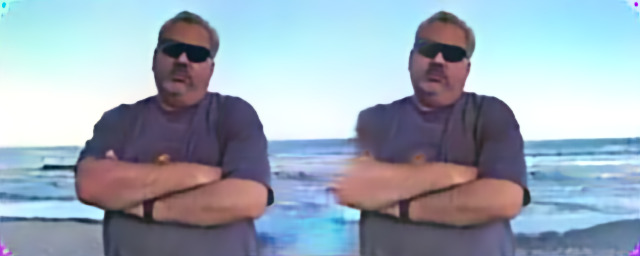}\caption{DCIL \cite{deep2019dcil}  }
\end{subfigure}
\begin{subfigure}[b]{0.26\linewidth}
\includegraphics[width=\linewidth, height=3cm,keepaspectratio]{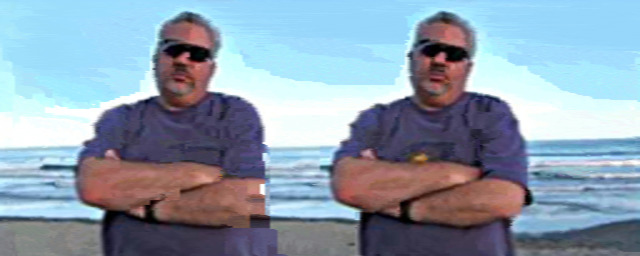}\caption{DeepCFL} 
\end{subfigure}} \end{center} \vspace*{-0.4cm}
\caption{The challenge is to synthesize a new object \cite{shocher2019inGan, deep2019dcil}. DeepCFL observed that object synthesis is achievable at a different scale, similar to DCIL\cite{deep2019dcil}. DeepCFL output image with better features near the elbow, but the background is not clear.
}\label{fig: retargetingGAN2}
\end{figure}%
\subsection{Image Resize}
We have discussed image outpainting, which is different from content-aware image resize, where the task is to resize the image while preserving the salient objects of the image \cite{shocher2019inGan}. DeepCFL is able to synthesize new objects when resizing the input image (Fig.~\ref{fig: retargetingGAN1}). The source image is scaled $2\times$ along the height and the width. Therefore, the pixel correspondence between the source and the generated target images is not well defined. The image resize is done by using the generator to scale the input and then computing the adversarial loss in a cycle consistent way. 

Fig.~\ref{fig: retargetingGAN2} show the challenging scenario of object synthesis for various single image GAN frameworks. Inspired by InGAN \cite{shocher2019inGan}, our framework DeepCFL studies deep contextual features. DeepCFL is different from DCIL \cite{deep2019dcil} as it uses the adversarial framework on VGG features space for image outpainting. In contrast, DCIL uses the adversarial framework on the image space for Denoising-super resolution. We believe that the results of various single image GAN framework in Fig.~\ref{fig: retargetingGAN2} could be improvised further.

\subsection{Restoration of $50\%$ pixels.} 
\label{ssec: restore}  
To investigate contextual features leaning in the presence of a high degree of corruption, we perform restoration of $50\%$ missing pixels spread across the entire image uniformly at random. It is a different setup than outpainting and inpainting, where one has to fill a missing region (\textit{i.e.}, a contiguous array of pixels). We further increase the task difficulty by using the corrupted image containing a word cloud. We denote the above setup as RestoreWC 50\% (WC denotes word-cloud). It is a challenging setup because the small font present in the corrupted image would require to fill fine image features details.

We show image restoration in RestoreWC 50\% setup in Fig.~\ref{fig: inpaintingWd}. The quantitative comparison (SSIM) for RestoreWC 50\% is as follows. DIP \cite{Ulyanov2018CVPR}: 0.92, MEDS \cite{mastan2019multi}: 0.93, InGAN \cite{shocher2019inGan}: 0.92, and DeepCFL (ours): 0.92. It could be observed that DeepCFL performs comparably to other frameworks.  It might be because the image features computed from the highly corrupted image might not be sufficient for restoration in the single image GAN framework. Therefore, contextual learning is a bit less effective. We believe that the pixel-based loss would not have the object synthesis abilities of the single image GAN frameworks (Fig.~\ref{fig: retargetingGAN2}). 



\begin{figure}[!ht]\begin{center} \captionsetup[sub]{font=footnotesize,labelfont={bf,sf}, justification=centering}   \captionsetup{font=small,labelfont={bf,sf}}\resizebox{\linewidth}{!}{
\begin{subfigure}{0.22\linewidth}\begin{center}
\includegraphics[width=\textwidth]{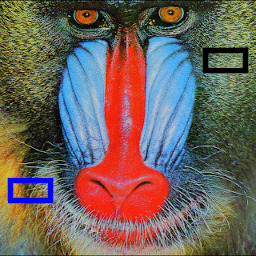} \\
\includegraphics[width=0.49\textwidth, cfbox=blue 0.5pt 0.5pt]{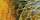}%
\includegraphics[width=0.49\textwidth, cfbox=black 0.5pt 0.5pt]{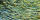}
\end{center} \vspace*{-0.2cm}\caption{Original \\image} \end{subfigure}
\begin{subfigure}{0.22\linewidth}\begin{center}
\includegraphics[width=\textwidth]{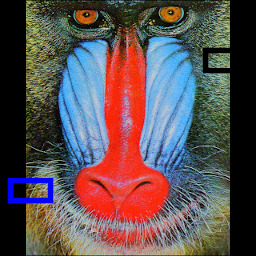} \\
\includegraphics[width=0.49\textwidth, cfbox=blue 0.5pt 0.5pt]{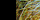}%
\includegraphics[width=0.49\textwidth, cfbox=black 0.5pt 0.5pt]{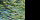}
\end{center} \vspace*{-0.2cm}\caption{Corrupted \\image} \end{subfigure}
\begin{subfigure}{0.22\linewidth}\begin{center}
\includegraphics[width=\textwidth]{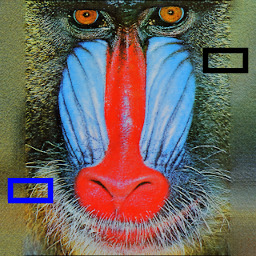} \\
\includegraphics[width=0.49\textwidth, cfbox=blue 0.5pt 0.5pt]{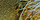}%
\includegraphics[width=0.49\textwidth, cfbox=black 0.5pt 0.5pt]{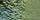}
\end{center} \vspace*{-0.2cm}\caption{DIP \cite{Ulyanov2018CVPR} \\ $\;$} \end{subfigure}
\begin{subfigure}{0.22\linewidth}\begin{center}
\includegraphics[width=\textwidth]{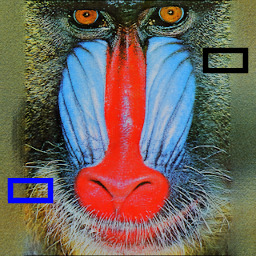} \\
\includegraphics[width=0.49\textwidth, cfbox=blue 0.5pt 0.5pt]{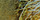}%
\includegraphics[width=0.49\textwidth, cfbox=black 0.5pt 0.5pt]{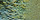}
\end{center} \vspace*{-0.2cm}\caption{DIP\cite{Ulyanov2018CVPR} \\ + CL} \end{subfigure}
\begin{subfigure}{0.22\linewidth}\begin{center}
\includegraphics[width=\textwidth]{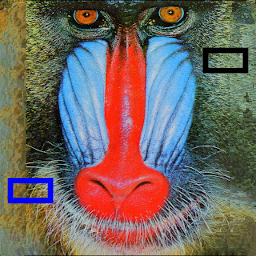} \\
\includegraphics[width=0.49\textwidth, cfbox=blue 0.5pt 0.5pt]{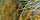}%
\includegraphics[width=0.49\textwidth, cfbox=black 0.5pt 0.5pt]{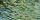}
\end{center} \vspace*{-0.2cm}\caption{DeepCFL \\ $\;$} \end{subfigure}}\end{center}\vspace*{-0.3cm} 
\caption{\small \textbf{Ablation study (1).} The figure shows the outpainting of 20\% pixels. DIP is a pixel-loss based setup. We integrated contextual loss (CL) with DIP \cite{Ulyanov2018CVPR} in ``DIP \cite{Ulyanov2018CVPR}  + CL" to show image restoration using CL and without GAN framework. DeepCFL is a GAN framework and it is observed to restore image features well.
}\label{fig: AblationContext}
\end{figure}%

\section{Ablation Studies and Limitations}\label{sec: ablationLimit} 
We show the usefulness of contextual learning in the adversarial framework in Fig.~\ref{fig: AblationContext}. The restored image features are highlighted in the cropped images. It could be observed that the single image GAN framework (DeepCFL) synthesizes image features for image restoration.

\begin{figure}[!ht] \captionsetup[sub]{font=small,labelfont={bf,sf}} \captionsetup{font=small,labelfont={bf,sf}} \centering \resizebox{\linewidth}{!}{
\begin{subfigure}{0.198\linewidth} \captionsetup{justification=centering} \begin{center}
\includegraphics[width=\textwidth]{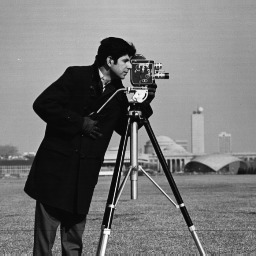}
\end{center}\caption{Original\\ image}
\end{subfigure} %
\begin{subfigure}{0.198\linewidth} \captionsetup{justification=centering} \begin{center}
\includegraphics[width=\textwidth]{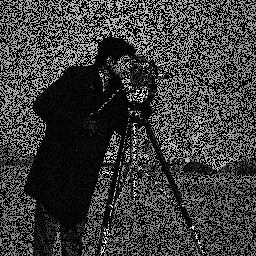}
\end{center}\caption{Masked\\ image}
\end{subfigure} %
\begin{subfigure}{0.198\linewidth} \captionsetup{justification=centering} \begin{center}
\includegraphics[width=\textwidth]{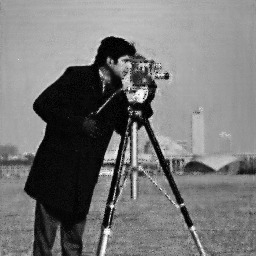}
\end{center}\caption{CAL,\\ 0.88}
\end{subfigure} %
\begin{subfigure}{0.198\linewidth} \captionsetup{justification=centering} \begin{center}
\includegraphics[width=\textwidth]{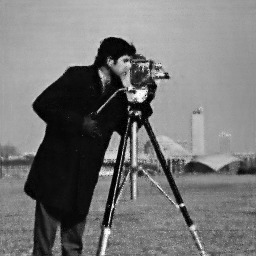}
\end{center}\caption{CVL,\\ 0.91}
\end{subfigure} %
\begin{subfigure}{0.198\linewidth} \captionsetup{justification=centering} \begin{center}
\includegraphics[width=\textwidth]{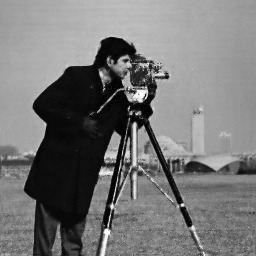}
\end{center}\caption{CFL,\\ 0.92}
\end{subfigure}} \vspace*{-0.2cm} %
\caption{\textbf{Ablation study (2).} The input is the masked image, which contains 50\% corrupted pixels. Here, the contextual feature loss CFL = CAL + CVL. It could be observed that CAL and CVL together enhance the restoration quality in CFL.}
\label{fig: ablation2}
\end{figure}

In Fig.~\ref{fig: ablation2}, we show an ablation study to disentangle the reconstruction using context vector loss (CVL), context adversarial loss (CAL), and contextual features loss (CFL) as defined in Sec.~\ref{ssec: lossFn}. The CFL setup performs better as it uses adversarial learning and context vector learning together. 

\begin{figure}[!ht] \captionsetup[sub]{font=footnotesize,labelfont={bf,sf}} \captionsetup{font=small,labelfont={bf,sf}} \resizebox{\linewidth}{!}{
\begin{subfigure}{0.22\linewidth}\begin{center}
\includegraphics[width=\textwidth]{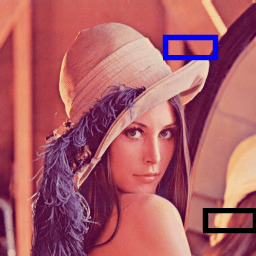} \\
\begin{subfigure}{0.48\textwidth} \includegraphics[width=0.95\textwidth, cfbox=blue 0.5pt 0.5pt]{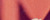} \end{subfigure}%
\begin{subfigure}{0.48\textwidth} \includegraphics[width=0.95\textwidth, cfbox=black 0.5pt 0.5pt]{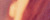} \end{subfigure} \caption{Original} \end{center} \end{subfigure} 
\begin{subfigure}{0.22\linewidth}\begin{center}
\includegraphics[width=\textwidth]{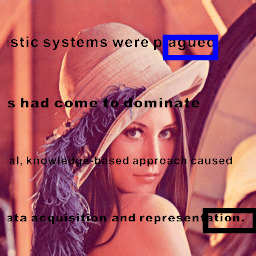} \\
\begin{subfigure}{0.48\textwidth} \includegraphics[width=0.95\textwidth, cfbox=blue 0.5pt 0.5pt]{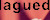}  \end{subfigure}%
\begin{subfigure}{0.48\textwidth} \includegraphics[width=0.95\textwidth, cfbox=black 0.5pt 0.5pt]{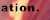} \end{subfigure} \caption{Masked} \end{center} \end{subfigure} 
\begin{subfigure}{0.22\linewidth}\begin{center}
\includegraphics[width=\textwidth]{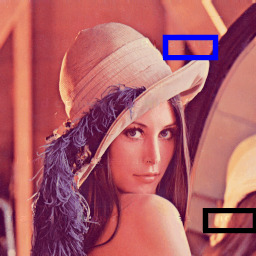} \\
\begin{subfigure}{0.48\textwidth} \includegraphics[width=0.95\textwidth, cfbox=blue 0.5pt 0.5pt]{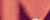}  \end{subfigure}%
\begin{subfigure}{0.48\textwidth} \includegraphics[width=0.95\textwidth, cfbox=black 0.5pt 0.5pt]{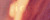} \end{subfigure} \caption{SSD} \end{center} \end{subfigure} 
\begin{subfigure}{0.22\linewidth}\begin{center}
\includegraphics[width=\textwidth]{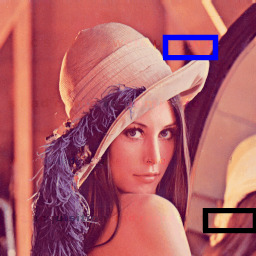} \\
\begin{subfigure}{0.48\textwidth} \includegraphics[width=0.95\textwidth, cfbox=blue 0.5pt 0.5pt]{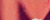}  \end{subfigure}%
\begin{subfigure}{0.48\textwidth} \includegraphics[width=0.95\textwidth, cfbox=black 0.5pt 0.5pt]{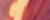} \end{subfigure}
 \caption{MSD} \end{center} \end{subfigure} } \vspace*{-0.2cm} 
\caption{\textbf{Ablation study (3).} The figure shows text removal in the presence of single-scale discriminator (SSD) and multi-scale discriminators (MSD) setups. SSD setup makes thin marks in the restored output, which are a bit less detectable in the MSD setup.}
\label{fig: AblationScale}
\end{figure}

Fig.~\ref{fig: AblationScale} shows the restoration in the presence of two discriminator architectures setup: single scale discriminator (SSD) and multiscale discriminator (MSD). InGAN \cite{shocher2019inGan} shows that MSD improves the performance significantly for image synthesis. We observed that higher model capacity did not significantly improve image restoration, similar to \cite{mastan2019multi} as the masked SSIM for SSD setup is (0.971) is close to  MSD setup (0.976). The visual performance enhancement would be because MSD setup enforces image statistics consistency at multiple levels, which is harder than solving at a single scale SSD setup. Our intuition is that solving a hard problem would help to learn better image features \cite{mastan2019multi}. Moreover, quantitative enhancement is close. Our interpretation of it is as follows. MSD in DeepCFL is operating on the context vectors. The scaling of the context vectors in MSD of DeepCFL and scaling the image in  \cite{shocher2019inGan, shaham2019singan, deep2019dcil} are completely different operations. The performance enhancement for image restoration using the scaling of context vector might not be very effective. 

\begin{figure}[!t] \centering \captionsetup[sub]{font=footnotesize,labelfont={bf,sf}} \captionsetup{font=small,labelfont={bf,sf}} \resizebox{\linewidth}{!}{
\begin{subfigure}{0.22\linewidth}\includegraphics[width=\linewidth]{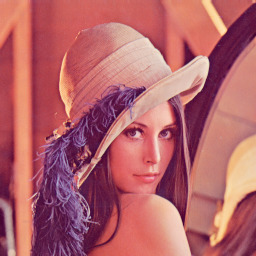} \caption{Original} \end{subfigure} 
\begin{subfigure}{0.22\linewidth}\includegraphics[width=\linewidth]{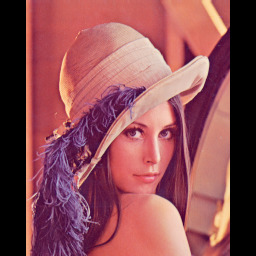} \caption{Masked} \end{subfigure} 
\begin{subfigure}{0.22\linewidth}\includegraphics[width=\linewidth]{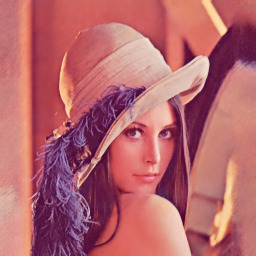} \caption{DIP \cite{Ulyanov2018CVPR} } \end{subfigure} 
\begin{subfigure}{0.22\linewidth}\includegraphics[width=\linewidth]{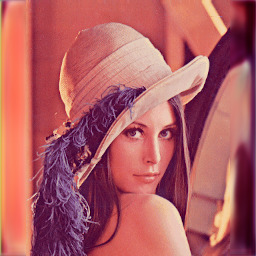} \caption{DeepCFL} \end{subfigure} 
}
\caption{\textbf{Limitation (1).} The aim is to restore a partially present object in the corrupted image (\textit{i.e.}, head). The features in the masked image is not enough to restore head in the mirror. Therefore, we could observe that the reconstruction using DIP \cite{Ulyanov2018CVPR} and DeepCFL is not performed well.}
\label{fig: failureEx}
\end{figure}

Fig.~\ref{fig: failureEx} shows the reconstruction when the information in the corrupted image is not sufficient to fill the missing regions.  The limitation is due to the lack of feature learning from the training samples in the single image GAN framework. A similar limitation has also been reported for image manipulation tasks \cite{shaham2019singan}.  Restoration of an object which is partially present in the image would also be exciting. However, it is not within the scope of this work.

\begin{figure}[!t] \captionsetup[sub]{font=footnotesize,labelfont={bf,sf}} \captionsetup{font=small,labelfont={bf,sf}} \centering \resizebox{\linewidth}{!}{ 
\begin{subfigure}{0.18\linewidth}\captionsetup{justification=centering}\includegraphics[width=\linewidth]{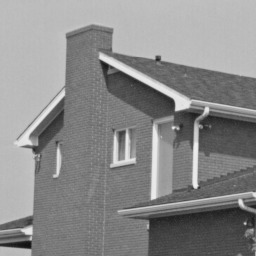} \caption{ Real \\image}
\end{subfigure}
\begin{subfigure}{0.18\linewidth}\captionsetup{justification=centering}\includegraphics[width=\linewidth]{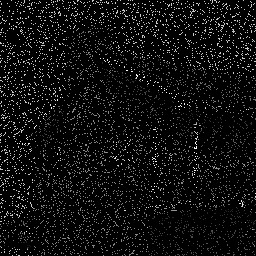} \caption{ Masked \\image}
\end{subfigure} 
\begin{subfigure}{0.18\linewidth}\captionsetup{justification=centering}\includegraphics[width=\linewidth]{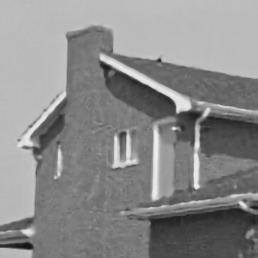} \caption{ DIP \\0.92 } 
\end{subfigure}
\begin{subfigure}{0.18\linewidth}\captionsetup{justification=centering}\includegraphics[width=\linewidth]{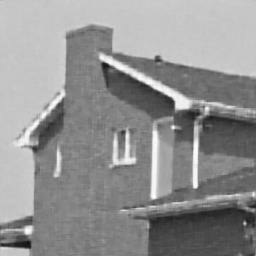} \caption{ MEDS \\0.91} 
\end{subfigure}
\begin{subfigure}{0.18\linewidth} \captionsetup{justification=centering}
\includegraphics[width=\linewidth]{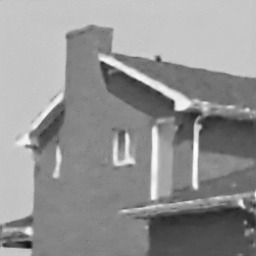} \caption{InGAN\\0.91} 
\end{subfigure}
\begin{subfigure}{0.18\linewidth}\captionsetup{justification=centering}\includegraphics[width=\linewidth]{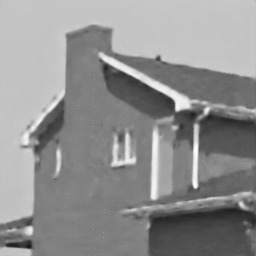} \caption{DeepCFL 0.91}
\end{subfigure} } \vspace*{-0.2cm} 
\caption{\textbf{Limitation (2).} The figure shows the restoration of 90\% pixels. DeepCFL preserves the image features comparable to DIP \cite{Ulyanov2018CVPR}, MEDS \cite{mastan2019multi}, and InGAN \cite{shocher2018internal}. } 
\label{fig: restorationBw}
\end{figure}

Fig.~\ref{fig: restorationBw} shows the the restoration of 90\% pixels ($r=90$) using image features learning from 10\% pixels. It could be observed that it is difficult to understand the semantics of the scene from 10\% pixels. The experiment confirms our observation that the adversarial learning of image context is less effective for the high degree of corruption. We show more results in the supplementary material.


\section{Discussion} 
DeepCFL is a single image GAN framework. The data-driven supervised feature learning setups use paired examples of ground truth (GT) and corrupted images. The corrupted images are fed into the network and generated outputs are matched with the GT image. DeepCFL is not trained by showing training samples of GT and corrupted images. DeepCFL can be fairly compared only with training data-independent methods as they also do not use training samples. Training based methods could synthesize image feature details that are not present in the input image, which is not possible in the training data-independent setups (Fig.~\ref{fig: failureEx} and Fig.~\ref{fig: restorationBw}). The feature extractor VGG-19 contains layers at different scales, where each layer contains varying levels of abstractions. We believe that combining features from various VGG-19 layers would be helpful. Moreover, it would increase the model complexity. The scope of DeepCFL is limited to the contextual features present in $conv4\_2$ layer. We propose as future work to perform studies on how to increase VGG19 layers for feature comparison while minimizing the computational overhead. 

\section{Conclusion}\label{sec: conclusion}
We investigate deep contextual features learning (CFL) in the single image GAN framework for image restoration and image synthesis. The main challenge to accomplish the above tasks is when the information contained in the input image is not sufficient for synthesizing the necessary image features. DeepCFL synthesizes image features based on the semantics to perform outpainting, inpainting, restoration of $r\%$ pixels, and image resizing. It would be interesting to study the performance of the single image GAN framework in the setting of videos similar to \cite{zhang2019internal, kim2019deep}.

{\small
\noindent \textbf{Acknowledgments.} Indra Deep Mastan was supported by Visvesvaraya Ph.D. fellowship. Shanmuganathan Raman was supported by SERB Core Research Grant and SERB MATRICS.
}

{\small
\bibliographystyle{ieee_fullname}
\bibliography{egbib}
}

\end{document}